\documentclass{article}

    \PassOptionsToPackage{numbers, compress}{natbib}

\usepackage[final]{neurips_2021}

\usepackage[utf8]{inputenc} 
\usepackage[T1]{fontenc}    
\usepackage{hyperref}       
\usepackage{url}            
\usepackage{booktabs}       
\usepackage{amsfonts}       
\usepackage{nicefrac}       
\usepackage{microtype}      
\usepackage{xcolor}         
\usepackage{makecell}
\usepackage{amsmath}
\usepackage{graphicx}

\usepackage{lipsum}   
\usepackage{amsthm}
\usepackage{amssymb}
\usepackage{algpseudocode}
\usepackage{algorithm}
\usepackage[toc,page]{appendix}

\newtheorem{proposition}{Proposition}

\usepackage{bm}
\graphicspath{ {figure/} }
\usepackage{epstopdf}
\usepackage{comment}
\usepackage{color}

\title{MixSeq: Connecting Macroscopic Time Series Forecasting with Microscopic Time Series Data}

\author{%
  Zhibo Zhu\thanks{Equal contribution.} \\
  Ant Group \\
  \texttt{gavin.zzb@antgroup.com} \\
  \And
  Ziqi Liu\footnotemark[1]\\
  Ant Group \\
  \texttt{ziqiliu@antgroup.com} \\
  \AND
  Ge Jin \\
  Ant Group \\
  \texttt{elvis.jg@antgroup.com}
  \And
  Zhiqiang Zhang \\
  Ant Group \\
  \texttt{lingyao.zzq@antgroup.com} \\
  \And
  Lei Chen \\
  Ant Group \\
  \texttt{qingli.cl@antgroup.com} \\
  \And
  Jun Zhou\thanks{Corresponding author.} \\
  Ant Group \\
  \texttt{jun.zhoujun@antgroup.com} \\
  \And
  Jianyong Zhou \\
  Ant Group \\
  \texttt{neil.zjy@antgroup.com} \\
}

\begin{document}
\maketitle

\begin{abstract}
Time series forecasting is widely used in business intelligence, e.g.,
forecast stock market price, sales, and help the analysis of
data trend. Most time series of interest are macroscopic time series
that are aggregated from microscopic data. However, instead of directly
modeling the macroscopic time series, rare literature studied the forecasting
of macroscopic time series by leveraging data on the microscopic level. In this paper,
we assume that the microscopic time series follow some unknown mixture
probabilistic distributions. We theoretically show that as we identify
the ground truth latent mixture components, the estimation of time series
from each component could be improved because of lower variance, thus
benefitting the estimation of macroscopic time series as well. Inspired
by the power of Seq2seq and its variants on the modeling of time series
data, we propose Mixture of Seq2seq (MixSeq), an end2end mixture model to
cluster microscopic time series, where all the components come from
a family of Seq2seq models parameterized by different parameters. 
Extensive experiments on both synthetic and real-world data
show the superiority of our approach.

\end{abstract}

\section{Introduction}
\label{sec:intro}
Time series forecasting has proven to be important to help people manage
resources and make decisions~\cite{lim2020time}. For example, probabilistic forecasting of
product demand and supply in retails~\cite{chen2019much}, or the forecasting of 
loans~\cite{abu1996introduction} in a financial institution can help people do 
inventory or financing planning to maximize the profit. Most time series of interest 
are macroscopic time series, e.g., the sales of an online retail platform, 
the loans of a financial institution, or the number of infections caused by 
some pandemic diseases in a state, that are comprised of microscopic time 
series, e.g., the sales of a merchant in the
online retail, the loans from a customer given the financial institution, or the
number of infections in a certain region. That is, the observed macroscopic time series
are just the aggregation or sum of microscopic time series.

Although various time series forecasting models, e.g., State Space Models (SSMs)~\cite{durbin2012time},
Autoregressive (AR) models~\cite{asteriou2011arima}, or deep neural networks~\cite{benidis2020neural}, 
have been widely studied for decades, all of them study the modeling of time series without considering
the connections between macroscopic time series of interest and the underlying 
time series on the microscopic level.

In this paper, we study the question whether the forecasting of macroscopic time 
series can be improved by leveraging the underlying microscopic time series, and the answer
is yes. Basically, though accurately modeling each microscopic time series could 
be challenging due to large variations, we show that by carefully clustering microscopic time series
into clusters, i.e., clustered time series, and using canonical approaches to model each of clusters,
finally we can achieve promising results by simply summing over the forecasting results of each cluster.

To be more specific,
\textbf{first}, we assume that the microscopic time
series are generated from a probabilistic mixture model~\cite{mclachlan1988mixture} 
where there exist $K$ components.
The generation of each microscopic time series is by first selecting a component $z$
from $\{1,...,K\}$ with a prior $p(z)$ (a Discrete distribution), then generating the microscopic
observation from a probabilistic distribution $p(x;\Phi_{z}, z)$ parameterized by 
the corresponding component $\Phi_z$. We show that
as we can identify the ground truth components of the mixture, and the ground
truth assignment of each microscopic observation, independent modeling of time series data
from each component could be improved due to lower variance, and further benefitting
the estimation of macroscopic time series that are of interest. 
\textbf{Second}, inspired by recent successes of Seq2seq 
models~\cite{vaswani2017attention,cho2014properties,du2018time} 
based on deep neural networks, e.g., variants
of recurrent neural networks (RNNs)~\cite{hewamalage2021recurrent,yu2017long,maddix2018deep}, 
convolutional neural networks (CNNs)~\cite{bai2018empirical,hao2020temporal}, and 
Transformers~\cite{li2019enhancing,wu2020deep}, 
we propose Mixture of Seq2seq (MixSeq), a mixture model for time series, where the 
components come from a family of Seq2seq models parameterized by different
parameters. \textbf{Third}, we conduct synthetic experiments to demonstrate the
superiority of our approach, and extensive experiments on real-world data to show 
the power of our approach compared with canonical approaches.

{\bfseries Our contributions}. We summarize our contributions in two-fold. \textbf{(1)} We
show that by transforming the original macroscopic time series via clustering,
the expected variance of each clustered time series could be optimized, thus improving the 
accuracy and robustness for the estimation of macroscopic time series. 
\textbf{(2)} We propose MixSeq which is an end2end mixture model with each component 
coming from a family of Seq2seq models. Our empirical results based on MixSeq 
show the superiority compared with canonical approaches.

\section{Background}
In this section, we first give a formal problem definition.
We then review the bases related to this work, and have
a discussion of related works.

{\bfseries Problem definition}. 
Let us assume a macroscopic time series $x_{1:t_0} = \left[x_{1},...,x_{t_0}\right]$,
and $x_t \in \mathbb{R}$ denotes the value of time series at time $t$. We aim to predict
the next $\tau$ time steps, i.e., $x_{t_0+1:t_0+\tau}$.
We are interested in the following conditional distribution
\begin{equation}\label{eq:problem}
p(x_{t_0+1:t_0+\tau}|x_{1:t_0}) = \prod_{t=t_0+1}^{t_0+\tau} p(x_t|x_{<t};\Theta),
\end{equation}
where $x_{<t}$ represents $x_{1:t-1}$ in interval $[1,t)$.
To study the above problem, we assume that the
macroscopic time series is comprised of $m$ microscopic time series,
i.e., $x_t = \sum_{i=1}^{m} x_{i,t}$ where $x_{i,t} \in \mathbb{R}$
denotes the value of the $i$-th microscopic time series at time $t$.
We aim to cluster the $m$ microscopic time series into $K$
clustered time series $\left\{ x_{1:t_0}^{(z)} \right\}_{z=1}^{K}$,
where $x_{t}^{(z)} = \sum_{\{i|z_i=z,\forall i\}} x_{i,t}$ given
the label assignment of the $i$-th microscopic time series 
$z_i \in \left\{ 1,...,K \right\}$. This is based on
our results in Section~\ref{sec:theory} that
the macroscopic time series forecasting can be improved
with optimal clustering. Hence, instead of directly modeling $p(x_{t_0+1:t_0+\tau})$,
we study the clustering of $m$ microscopic time series in Section~\ref{sec:mixseq},
and model the conditional distribution of clustered time series
$\left\{ p(x_{t_0+1:t_0+\tau}^{(z)}) \right\}_{z=1}^{K}$
with canonical approaches.

\subsection{Seq2seq: encoder-decoder architectures for time series}
\label{sec:seq2seq}
An encoder-decoder based neural network models
the conditional distribution Eq.~\eqref{eq:problem} as a distribution
from the exponential families, e.g., Gaussian, Gamma or Binomial distributions, 
with sufficient statistics generated from a neural network.
The encoder feeds $x_{<t}$ into a neural architecture, e.g., RNNs, CNNs
or self-attentions, to generate the representation of historical time series,
denoted as $h_{t}$, then we use a decoder to yield the result $x_{t}$. After $\tau$ 
iterations in an autoregressive style, it finally generates the whole time series to be predicted.

To instantiate above Seq2seq architecture, we denote $o_{1:t}$, where $o_t \in \mathbb{R}^d$, as covariates that 
are known a priori, e.g., dates. We denote $Y_{t} = \left[ x_{1:t-1} \,\,\|\,\, o_{2:t} 
\right] \in \mathbb{R}^{(t-1)\times (d+1)}$ where we use $\|$ for concatenation.
The encoder generates the representation $h_{t}$ of $x_{<t}$ via 
Transformer~\cite{vaswani2017attention,li2019enhancing} as follows. 
We first transform $Y_{t}$ by some functions $\rho(\cdot)$, e.g., causal convolution~\cite{li2019enhancing}
to $H^{(0)} = \rho(Y_{t}) \in \mathbb{R}^{(t-1)\times d_k}$. 
Transformer then iterates the following self-attention layer $L$ times:
\begin{equation}\label{eq:transformer}
\begin{aligned}
&H^{(l)} = \mathrm{MLP}^{(l)}(H^{(\mathrm{tmp})}),\,\,
H^{(\mathrm{tmp})} = \mathrm{SOFTMAX}\left( \frac{Q^{(l)} K^{(l)\top}}{\sqrt{d_q}} M \right) V^{(l)},\\
&Q^{(l)} = H^{(l-1)}W_q^{(l)}, K^{(l)} = H^{(l-1)}W_k^{(l)}, V^{(l)} = H^{(l-1)}W_v^{(l)}.
\end{aligned}
\end{equation}
That is, we first transform $Y$\footnote{We ignore the subscript 
for simplicity in condition that the context is of clarity.} into query, key, and value 
matrices, i.e., $Q = Y W_q$, $K = Y W_k$, and $V = Y W_v$
respectively, where $W_q \in \mathbb{R}^{d_k\times d_q}, 
W_k \in \mathbb{R}^{d_k\times d_q}, W_v \in \mathbb{R}^{d_k\times d_v}$ in each layer are
learnable parameters. Then we do scaled inner product attention to yield $H^{(l)} \in \mathbb{R}^{(t-1)\times d_k}$ where
$M$ is a mask matrix to filter out rightward attention by setting all 
upper triangular elements to $-\infty$. We denote $\mathrm{MLP}(\cdot)$ as a multi-layer perceptron function.
Afterwards, we can generate the representation $h_{t} \in \mathbb{R}^{d_p}$ for $x_{<t}$
via $h_{t} = \nu(H^{(L)})$ where we denote $\nu(\cdot)$ as a deep set function~\cite{zaheer2017deep}
that operates on rows of $H^{(L)}$, i.e., $\nu(\{H_1^{(L)},..., H_{t-1}^{(L)}\})$. We denote
the feedforward function to generate $H^{(L)}$ as $H^{(L)} \sim g(H^{(0)})$, i.e., Eq~\eqref{eq:transformer}.

Given $h_{t}$, the decoder generates the sufficient statistics and finally yields 
$x_{t} \sim p(x;\mathrm{MLP}(h_{t}))$ from a distribution in the exponential family.

\subsection{Related works}
{\bfseries Time series forecasting} has been studied for decades. 
We summarize works related to time series forecasting into two categories.
\textbf{First}, many models come from the family of autoregressive 
integrated moving average (ARIMA)~\cite{box1968some,asteriou2011arima}, where AR 
indicates that the evolving variable of interest is regressed on 
its own lagged values, the MA indicates that the regression error 
is actually a linear combination of error terms, and the ``I'' 
indicates that the data values have been replaced with the 
difference between their values and the previous values to handle 
non-stationary~\cite{pemberton1990non}. The State Space Models 
(SSM)~\cite{durbin2012time} aim
to use state transition function to model the transfer of states and generate
observations via a observation function. These statistical approaches typically model
time series independently, and most of them only utilize values from history
but ignore covariates that are important signals for forecasting.
\textbf{Second}, as rapid development of deep neural networks, people started 
studying many neural networks for the modeling of time series~\cite{benidis2020neural,lim2020time}. 
Most successful neural networks are based on the encoder-decoder 
architectures~\cite{vaswani2017attention,cho2014properties,du2018time,sutskever2014sequence,bahdanau2014neural,dama2021analysis,lim2020time,ma2019learning}, namely Seq2seq. Basically, various Seq2seq models based 
on RNNs~\cite{hewamalage2021recurrent,salinas2020deepar,wen2017multi,lai2018modeling,yu2017long,maddix2018deep}, 
CNNs~\cite{bai2018empirical,hao2020temporal}, and 
Transformers (self-attentions)~\cite{li2019enhancing,wu2020deep}
are proposed to model the non-linearity for time series.

No matter models studied in statistics or deep neural networks,
these works mainly focus on the forecasting of single or multivariate
time series, but ignore the auxiliary information that the time series could be
made up of microscopic data.

{\bfseries Time series clustering} is another topic for exploratory analysis
of time series. We summarize the literature into three categories, i.e., 
study of distance functions, generative models, and feature extraction 
for time series. \textbf{First}, Dynamic time wrapping~\cite{petitjean2011global}, 
similarity metric that measures temporal dynamics~\cite{yang2011patterns}, and specific 
measures for the shape~\cite{paparrizos2015k} of time series are proposed 
to adapt to various time series characteristics, e.g., scaling and distortion.
Typically these distance functions are mostly manually defined and cannot generalize
to more general settings. \textbf{Second}, generative model based approaches assume 
that the observed time series is generated by an underlying model, such as 
hidden markov model~\cite{oates1999clustering} or mixture of ARMA~\cite{xiong2004time}. 
\textbf{Third}, early studies on feature extraction of time series are based on component 
analysis~\cite{guo2008time}, and kernels, e.g., 
u-shapelet~\cite{zakaria2012clustering}. As the development of deep neural networks,
several encoder-decoder architectures~\cite{madiraju2018deep,ma2019learning} are proposed to 
learn better representations of time series for clustering. 

However, the main purpose of works in this line is to conduct exploratory analysis
of time series, while their usage for time series forecasting has never been studied.
That is, these works define various metrics to evaluate the goodness of the clustering
results, but how to learn the optimal clustering for time series forecasting remains
an open question.

\section{Microscopic time series under mixture model}
\label{sec:theory}
We analyze the variance of mixture models, and further verify
our results with simple toy examples.

\subsection{Analyses on the variance of mixture model}
In this part, we analyze the variance of probabilistic mixture models.
A mixture model~\cite{mclachlan1988mixture} is a probabilistic model for representing 
the presence of subpopulations within an overall population.
Mixture model typically consists of a prior that represents
the probability over subpopulations, and components, each of which
defines the probability distribution of the corresponding subpopulation.
Formally, we can write
\begin{equation}\label{eq:mixture}
f(x) = \sum_i p_i \cdot f_i(x),
\end{equation}
where $f(\cdot)$ denotes the mixture distribution, $p_i$ denotes
the prior over subpopulations, and $f_i(\cdot)$ represents the distribution
corresponding to the $i$-th component.


\begin{proposition}
Assuming the mixture model with probability density function $f(x)$, and corrsponding components $\left\{f_i(x)\right\}_{i=1}^K$ with constants $\left\{p_i\right\}_{i=1}^K$ ($\left\{p_i\right\}_{i=1}^K$ lie in a simplex), we have $f(x) = \sum_i p_i f_i(x)$.
In condition that $f(\cdot)$ and $\left\{f_i(\cdot)\right\}_{i=1}^K$ have first and second moments, i.e., $\mu^{(1)}$ and $\mu^{(2)}$ for $f(x)$, 
and $\left\{\mu_i^{(1)}\right\}_{i=1}^K$ and $\left\{\mu_i^{(2)}\right\}_{i=1}^K$ for components $\left\{f_i(x)\right\}_{i=1}^K$, we have:
\begin{align}
\sum_i p_i\cdot \mathrm{Var}(f_i) \leq \mathrm{Var}(f).
\end{align}
\end{proposition}
We use the fact that $\mu^{(k)} = \sum_i p_i \mu_i^{(k)}$. By using Jensen's Inequality 
on $\sum_i p_i \left( \mu_i^{(1)} \right)^2 \geq 
\left( \sum_i p_i \mu_i^{(1)} \right)^2$, we immediately yield the result. See detailed 
proofs in supplementary.


This proposition states that, if we have limited data samples 
(always the truth in reality) and in case we know the ground 
truth data generative process a priori, i.e., the exact generative 
process of each sample from its corresponding component, the variance on
expectation conditioned on the ground truth data assignment
should be no larger than the variance of the mixture. Based on the
assumption that microscopic data are independent, the variance of the
aggregation of clustered data should be at least no larger than the
aggregation of all microscopic data, i.e., the macroscopic data. So the 
modeling of clustered data from separate components could possibly be more
accurate and robust compared with the modeling of macroscopic data.
This result motivates us to forecast macroscopic time series by clustering 
the underlying microscopic time series. Essentially, we transform the original 
macroscopic time series data to clusters with lower variances using a 
clustering approach, then followed by any time series models to forecast 
each clustered time series. After that, we sum over all the results from 
those clusters so as to yield the forecasting of macroscopic time series. 
We demonstrate this result with toy examples next.

\subsection{Demonstration with toy examples}
We demonstrate the effectiveness of forecasting macroscopic time series by 
aggregating the forecasting results from clustered time series.

{\bfseries Simulation setting.}
We generate microscopic time series from a mixture model, 
such as Gaussian process (GP)~\cite{roberts2013gaussian} or ARMA~\cite{box2015time} with $3$ or $5$ components. 
We generate $5$ time series for each component, and yield $15$ or $25$ microscopic time 
series in total. We sum all the time series as the macroscopic time series. We get
clustered time series by simply summing microscopic time series from the same component. 
Our purpose is to compare the performance between forecasting results directly on macroscopic time 
series (macro results) and sum of forecasting results of clustered time series (clustered results). 
We set the length of time series as $360$, and use rolling window approach for training and validating 
our results in the last $120$ time steps (i.e., at each time step, we train the model 
using the time series before current time point, and validate using the following 
$30$ values). We fit the data with either GP or ARMA depending on the generative model. 
We describe the detailed simulation parameters of mixture models in supplementary.

{\bfseries Simulation results.}
Table~\ref{tab:toy} shows the results measured by symmetric mean absolute percentage error (SMAPE)\footnote{Details are in supplementary}. It is obvious that no matter time series generated by mixture of GP or mixure of ARMA, the clustered results are superior to macro results. In other words, if we knew the ground truth component of each microscopic time series, modeling on clustered data aggregated by time series from the same component would have better results compared with directly modeling the macroscopic time series.

\begin{table}
  \caption{We run the experiments $5$ times, and show the average results (SMAPE) of macro results and clustered results with ground truth clusters. Lower is better.}
  \label{tab:toy}
  \centering
  \begin{tabular}{lllll}
    \toprule
    & \multicolumn{2}{c}{GP time series data} & \multicolumn{2}{c}{ARMA time series data} \\
    \cmidrule(r){2-3} \cmidrule(r){4-5}
    & 3 clusters & 5 clusters & 3 clusters & 5 clusters \\
    \midrule
    macro results & 0.0263 & 0.0242 & 0.5870 & 0.5940 \\ 
    clustered results & {\bf 0.0210} & {\bf 0.0198} & {\bf 0.3590} & {\bf 0.3840} \\ 
    \bottomrule
  \end{tabular}
\end{table}
  

\section{MixSeq: a mixture model for time series}
\label{sec:mixseq}
Based on the analysis in Section~\ref{sec:theory}, we assume 
microscopic time series follow a mixture distribution, and propose
a mixture model, MixSeq, to cluster microscopic time series. 


\subsection{Our model}
Our model assumes that each of $m$ microscopic time series follows a mixture
probabilistic distribution. To forecast $x_{t_0+1:t_0+\tau}$ given $x_{1:t_0}$,
our approach is to first partition $\left\{x_{i,1:t_0}\right\}_{i=1}^m$ into
$K$ clusters via MixSeq. Since
the distribution is applicable to all microscopic time series, we ignore
the subscript $i$ and time interval $1:t_0$ for simplicity. 
We study the following generative probability of $x \in \mathbb{R}^{t_0}$:
\begin{equation}\label{eq:mix-model}
p(x) = \sum_z p(x,z) = \sum_z p(z)p(x|z) = \sum_z p(z) \prod_{t=1}^{t_0} 
p(x_t|x_{<t},z; \Phi_z),
\end{equation}
where $z \in \{1,2,...,K\}$ is the discrete latent variable, 
$K$ is the number of components in the mixture model, 
$p(z)$ is the prior of cluster indexed by $z$, and $p(x|z)$ is the probability of time series $x$ generated 
by the corresponding component governed by parameter $\Phi_z$. Note that
we have $K$ parameters $\Theta = \{\Phi_1, \Phi_2, \dots, \Phi_K\}$ in the mixture model.

We use ConvTrans introduced in~\cite{li2019enhancing} as our backbone 
to model the conditional $p(x_t|x_{<t},z; \Phi_z)$. 
To instantiate, we model the conditional $p(x_t|x_{<t},z; \Phi_z)$ 
by first generating $H^{(0)}\sim \rho(Y_t)$ via causal convolution~\cite{li2019enhancing}. 
We then generate $H^{(L)} \sim g(H^{(0)}) \in \mathbb{R}^{(t-1) \times d_k}$ given $x_{<t}$. 
Finally we generate the representation for $x_{<t}$ as 
$h_{t} = \nu(H^{(L)}) = \sigma(W_s \sum_{j=1}^{t-1} H_j^{(L)}) \in \mathbb{R}^{d_p}$, 
where $W_s \in \mathbb{R}^{d_p\times d_k}$ and $\sigma$ as ReLU activation function. Afterwards, we 
decode $h_{t}$ to form the specific distribution from an exponential family.
In particular, we use Gaussian distribution $p(x_t|x_{<t},z; \Phi_z) = \mathcal{N}(x_t;\mu_t, \sigma_t^2)$, 
where the mean and variance can be generated by following transformations,
\begin{equation}
  \begin{aligned}
    \mu_t = w_{\mu}^T h_t + b_{\mu},\,\,\,\sigma_t^2 = \mathrm{log}(1+\mathrm{exp}(w_{\sigma}^T h_t + b_{\sigma})),
  \end{aligned}
\end{equation}
where $w_{\mu},w_{\sigma} \in \mathbb{R}^{d_p}$ are parameters, and $b_{\mu},b_{\sigma} \in \mathbb{R}$ are biases.

\subsection{Posterior inference and learning algorithms}
We aim to learn the parameter $\Theta$ and efficiently
infer the posterior distribution of $p(z|x)$ in Eq.~\eqref{eq:mix-model}.
However, it is intractable to maximize the marginal likelihood $p(x)$ after
taking logarithm, i.e., $\log p(x)$, since of the involvement of logarithm of sum. 
To tackle this non-convex problem, we resort to stochastic auto-encoding variational Bayesian 
algorithm (AEVB)~\cite{kingma2013auto}.
Regarding single microscopic time series, the variational lower bound (LB)~\cite{kingma2013auto} on 
the marginal likelihood is as below.
\begin{equation}\label{lb}
  \begin{aligned}
    \log p(x) &= \log \sum_z p(x,z) \ge \sum_z q(z|x) \log \frac{p(x,z)}{q(z|x)} \\
    &=\mathbb{E}_{q(z|x)} \log p(x|z) - \mathrm{KL}\left(q(z|x)\|p(z)\right) = \mathrm{LB},
  \end{aligned}
\end{equation}
where $q(z|x)$ is the approximated posterior of the latent variable $z$ given time series $x$. 
The benefit of using AEVB is that we can treat $q(z|x)$ as an encoder modeled by a neural network.
Hence, we reuse the ConvTrans~\cite{li2019enhancing} as our backbone, 
and model $q(z|x)$ as:
\begin{equation}
\begin{aligned}
q(z|x) = \mathrm{SOFTMAX}(W_a \cdot \nu(H^{(L)})),\,\,\,\, H^{(L)} = g(\rho(Y_{t_0})),
\end{aligned}
\end{equation}
where we denote $Y_{t_0} = [x_{1:t_0} \| o_{1:t_0}] \in \mathbb{R}^{t_0 \times (d+1)}$, 
$\nu(H^{(L)}) = \sigma(W_s\cdot \sum_{j=1}^{t_0} H_j^{(L)})$
with parameter $W_s \in \mathbb{R}^{d_p\times d_k}$ is the deep set function, 
and $W_a \in \mathbb{R}^{K\times d_p}$ as 
parameters to project the encoding to $K$ dimension. After the softmax operator, we 
derive the posterior distribution that lies in a simplex of $K$ dimension. 
Note that we use distinct $\rho(\cdot)$'s, $g(\cdot)$'s and $\nu(\cdot)$'s with different parameters to model 
$q(z|x)$ and $\{p(x_t|x_{<t}, z)\}_{z=1}^{K}$ respectively.
We assign each microscopic $x_i$ to cluster $z_i = \underset{z}{\arg\max}\ q(z|x_i)$ in our experiments.


{\bfseries Mode collapsing.} We find that directly optimizing the lower bound in 
Eq. (\ref{lb}) suffers from the mode collapsing problem. That is, the encoder 
$q(z|x)$ tends to assign all microscopic time series to one cluster, and does 
not effectively distinguish the data as expected, thus implying $I(x;z)=0$ ($I(\cdot)$ for mutual information).
In order to address the above mode collapsing problem, we add $I(x;z)$ to the lower 
bound in Eq. (\ref{lb}) which expects that the latent variable $z$ can extract discriminative 
information from different time series \cite{zhao2018unsupervised}. Then, we have
\begin{equation}
  \label{old-loss}
  \begin{aligned}
    &\mathbb{E}_x(\mathbb{E}_{q(z|x)}\log p(x|z))-\mathbb{E}_x(\mathrm{KL}(q(z|x)\|p(z)))+I(x;z)\\
    & = \mathbb{E}_x(\mathbb{E}_{q(z|x)}\log p(x|z))-\mathrm{KL}(q(z)\|p(z)),
    \end{aligned}
\end{equation}
where $q(z) = \frac{1}{m} \sum_{i=1}^{m} q(z|x_i)$ is an average of approximated posteriors over all microscopic data. 
We approximate this term by using a mini-batch of $m'$ samples, i.e., $q(z) = \frac{1}{m'} \sum_{i' \in \mathcal{B}} q(z|x_{i'})$.

{\bfseries Annealing tricks.}
Regarding long-length time series, the reconstruction loss and KL divergence in Eq. (\ref{old-loss}) 
are out of proportion. In this situation, the KL divergence has few effects on the optimization 
objective. So we finally derive the following objective to maximize:
\begin{equation}
  \label{new-loss}
  \mathbb{E}_x(\mathbb{E}_{q(z|x)}\log p(x|z))-\alpha\cdot \mathrm{KL}(q(z)\|p(z)) -\lambda\cdot \|\Theta \|
\end{equation}
where $\alpha$ is the trade-off hyperparameter. We use the following 
annealing strategy $\alpha = \mathrm{max}(a, b \times e^{(- \beta n)})$ to 
dynamically adjust $\alpha$ in the training process, where $\beta$ is the 
parameter controlling the rate of descent. Meanwhile, we also involve the $\ell_2$-norm 
regularizers on Seq2seq's parameters $\Theta$ with hyperparameter $\lambda\ge 0$.

\begin{table}
  \caption{Mean and standard deviation (SD, in bracket) of Rand Index (RI, the higher the better) by clustering on synthetic data generated by ARMA and DeepAR. MixSeq-infer represents that we infer the cluster of new data generated by different models after training MixSeq. On ARMA data, MixSeq and MixARMA have comparable performance; on DeepAR data, MixARMA degrades
  significantly which shows the effectiveness of MixSeq.}
  \label{tab-syndatares}
  \centering
  \begin{tabular}{lllll}
    \toprule
    & \multicolumn{2}{c}{ARMA synthetic data} & \multicolumn{2}{c}{DeepAR synthetic data} \\
    \cmidrule(r){2-3} \cmidrule(r){4-5}
    & 2 clusters & 3 clusters & 2 clusters & 3 clusters \\
    \midrule
    MixARMA & {\bf 0.9982}(0.0001) & 0.9509(0.1080) & 0.7995(0.2734) & 0.7687(0.0226) \\
    MixSeq & 0.9915(0.0024) & {\bf 0.9540}(0.0974) & {\bf 0.9986}(0.0003) & {\bf 0.8460}(0.0774) \\
    MixSeq-infer & 0.9929(0.0027) & 0.9544(0.0975) & 0.9982(0.0006) & 0.8460(0.0775) \\
    \bottomrule
  \end{tabular}
\end{table}

\section{Experimental results}\label{sec:exp}
We conduct extensive experiments to show the advantage of MixSeq. 
We evaluate the clustering performance of MixSeq on synthetic data,
present the results of macroscopic time series forecasting on real-world data,
and analyze the sensitivity of the cluster number of MixSeq.

\subsection{Synthetic datasets}
\label{exp-part1}
To demonstrate MixSeq's capability of clustering microscopic time 
series that follow various probabilistic mixture distributions, we conduct 
clustering experiments on synthetic data with ground truth. 
We generate two kinds of synthetic time series by ARMA~\cite{box2015time} and DeepAR~\cite{salinas2020deepar} respectively. 
For each model, we experiment with different number of clusters (2 and 3) 
generated with components governed by different parameters.

{\bfseries Experiment setting.} 
To generate data from ARMA, we use ARMA(2, 0) and 
$x_t = \phi_1 x_{t-1} + \phi_2 x_{t-2} + \epsilon_t$ 
with $\epsilon_t \sim N(0,0.27)$. We set parameters $[\phi_1, \phi_2]$ for three
components as $[-0.25, 0.52]$, $[0.34, 0.27]$, and $[1.5, -0.75]$
respectively. The synthetic time series from a mixture of $2$ components 
are generated using the first two components. The synthetic time series
from a mixture of $3$ components are generated using all $3$ components. 
To generate data from DeepAR, we use the DeepAR model with one LSTM layer, 
and the hidden number of units is $16$. Since it is difficult to 
randomly initialize the parameters of DeepAR, we train a base model on the 
real-world Wiki dataset~\cite{tran2021radflow} (discussed in section~\ref{exp-part2}). 
To build the other two DeepAR components, we respectively add random 
disturbance $\mathcal{N}(0,0.01)$ to the parameters of the base model. 
For each cluster, we generate $10,000$ time series with random initialized
sequences, and set the length of time series as $100$.

We use 1-layer causal convolution Transformer (ConvTrans \cite{li2019enhancing})
as our backbone model in MixSeq. We use the following parameters unless otherwise stated.
We set the number of multi-heads as $2$, kernel size as $3$, the number of kernel for causal convolution $d_k=16$, dropout rate as $0.1$, the penalty weight on the $\ell_2$-norm regularizer as 1e-5, and $d_p=d_v=16$. Meanwhile, we set the prior $p(z)$ as $1/K, \, \forall z$.
For the training parameters, we set the learning rate as 1e-4, batch size as 
$256$ and epochs as $100$. Furthermore, the $\alpha$ in MixSeq is annealed 
using the schedule $\alpha = \mathrm{max}(5, 20e^{(-0.03n)})$, where 
$n$ denotes the current epoch, and $\alpha$ is updated in the $[10,30,50]$-th epochs. 
For comparison, we employ MixARMA~\cite{xiong2004time}, a mixture of 
ARMA(2, 0) model optimized by EM algorithm~\cite{blei2017variational}, 
as our baseline. Both methods are evaluated using Rand Index (RI)~\cite{rand1971objective} 
(more details in supplementary).

{\bfseries Experiment results.}
We show the clustering performance of MixSeq and MixARMA on the synthetic data
in Table~\ref{tab-syndatares}. The results are given by the average of 5 
trials. Regarding the synthetic data from ARMA, both MixSeq and MixARMA perform very
well. However, for the synthetic data from DeepAR, MixARMA degrades significantly
while MixSeq achieves much better performance. This suggests that MixSeq can 
capture the complex nonlinear characteristics of time series generated by 
DeepAR when MixARMA fails to do so. Furthermore, we also generate new time 
series by the corresponding ARMA and DeepAR models, and infer their clusters 
with the trained MixSeq model. The performance is comparable with the 
training performance, which demonstrates that MixSeq actually captures the 
generation mode of time series.

\subsection{Real-world datasets}
\label{exp-part2}
We further evaluate the effectiveness of our model on the macroscopic time series
forecasting task. We compare MixSeq with existing clustering methods and
state-of-the-art time series forecasting approaches on several real-world
datasets. Specifically, for each dataset, the goal is to forecast the
macroscopic time series aggregated by all microscopic data. We cluster
microscopic time series into groups, and aggregate the time series in each
group to form the clustered time series. Then, we train the forecasting models
on the clustered time series separately, and give predictions of each clustered
time series. Finally, the estimation of macroscopic time series is obtained by 
aggregating all the predictions of clustered time series.

We report results on three real-world datasets, including 
Rossmann\footnote{https://www.kaggle.com/c/rossmann-store-sales}, 
M5\footnote{https://www.kaggle.com/c/m5-forecasting-accuracy} and 
Wiki~\cite{tran2021radflow}. The Rossmann dataset consists of historical 
sales data of $1,115$ Rossmann stores recorded every day. 
Similarly, the M5 dataset consists of $30,490$ microscopic time series as 
the daily sales of different products in ten Walmart stores in USA. 
The Wiki dataset contains $309,765$ microscopic time series 
representing the number of daily views of different Wikipedia articles. 
The dataset summary is shown in Table \ref{tab-realdata}, together with 
the setting of data splits.

\begin{table}
  \caption{Real-world dataset summary.}
  \label{tab-realdata}
  \centering
  \begin{tabular}{lllll}
    \toprule
    dataset & \makecell[l]{\# microscopic\\time series} & \makecell[l]{length of\\time series} & train interval & test internal \\
    \midrule
    Rossmann & 1115 & 942 & 20130101-20141231 & 20150101-20150731 \\
    M5 & 30490 & 1941 & 20110129-20160101 & 20160101-20160619 \\
    Wiki & 309765 & 1827 & 20150701-20191231 & 20200101-20200630 \\
    \bottomrule
  \end{tabular}
\end{table}

\begin{table}
  \caption{Comparisons on the microscopic time series clustering methods for macroscopic time series forecasting combined with three network-based forecasting methods: testing $R_{0.5}$/$R_{0.9}$-loss on three real-world datasets. Lower is better.}
  \label{tab-realdatarloss}
  \centering
  \begin{tabular}{llllll}
    \toprule
    & & Macro & DTCR & MixARMA & MixSeq \\
    \midrule
    & DeepAR & 0.1904/{\bf 0.0869} & 0.2292/0.1432 & 0.1981/0.1300 & {\bf 0.1857}/0.0987 \\
    Rossmann & TCN & 0.1866/0.1005 & 0.2023/0.1633 & 0.1861/0.1160 & {\bf 0.1728/0.0997} \\
    & ConvTrans & 0.1861/0.0822 & 0.2077/0.0930 & 0.1866/0.0854 &  {\bf 0.1847/0.0813} \\
    \midrule
    & DeepAR & {\bf 0.0548/0.0289} & 0.0787/0.0627 & 0.0624/0.0582 & 0.0582/0.0445 \\
    M5 & TCN & 0.0790/0.0635 & 0.0847/0.0805 & 0.0762/0.0789 & {\bf 0.0694/0.0508} \\
    & ConvTrans & 0.0553/0.0260 & 0.0514/0.0260 & 0.0497/0.0257 & {\bf 0.0460/0.0238} \\
    \midrule
    & DeepAR & 0.0958/0.0962 & 0.1073/0.1336 & 0.0974/0.1070 & {\bf 0.0939/0.0901} \\
    Wiki & TCN & 0.0966/0.1064 & 0.1237/0.1480 & 0.0963/0.1218 & {\bf 0.0886/0.0980} \\
    & ConvTrans & 0.0968/0.0589 & 0.1029/0.0531 & 0.0961/0.0594 & {\bf 0.0901/0.0516} \\
    \bottomrule
  \end{tabular}
\end{table}

{\bfseries Experiment setting.}
We summarize the clustering strategies for macroscopic time series 
forecasting as follows. \textbf{(1)} ``DTCR''~\cite{ma2019learning} 
is the deep temporal clustering representation method which integrates 
the temporal reconstruction, K-means objective and auxiliary classification 
task into a single Seq2seq model. \textbf{(2)} ``MixARMA''~\cite{xiong2004time} 
is the mixture of ARMA model that uses ARMA to capture the characteristics 
of microscopic time series. \textbf{(3)} ``MixSeq'' is our model with 1-layer 
causal convolution Transformer~\cite{li2019enhancing}. \textbf{(4)} We also 
report the results that we directly build forecasting model on the 
macroscopic time series without leveraging the microscopic data, 
named as ``Macro''.


For time series forecasting, we implement five methods combined with 
each clustering strategy, including ARMA~\cite{box2015time}, 
Prophet~\cite{taylor2018forecasting}, DeepAR~\cite{salinas2020deepar}, 
TCN~\cite{bai2018empirical}, and ConvTrans~\cite{li2019enhancing}. 
ARMA and Prophet give the prediction of point-wise value for time series, 
while DeepAR, TCN and ConvTrans are methods based on neural network for probabilistic 
forecasting with Gaussian distribution. We use the rolling window strategy on 
the test interval, and compare different methods in terms of the long-term 
forecasting performance for $30$ days. The data of last two months in train interval 
are used as validation data to find the optimal model.


We do grid search for the following hyperparameters in clustering and forecasting 
algorithms, i.e., the number of clusters $\{3,5,7\}$, the learning rate 
$\{0.001, 0.0001\}$, the penalty weight on the $\ell_2$-norm regularizers 
$\{1e-5,5e-5\}$, and the dropout rate $\{0,0.1\}$. The model with best 
validation performance is applied for obtaining the results on test interval. 
Meanwhile, we set batch size as $128$, and the number of training epochs as $300$ 
for Rossmann, $50$ for M5 and $20$ for Wiki. For DTCR, we use the same setting as~\cite{ma2019learning}.

Regarding time series forecasting models, we apply the default setting to ARMA and 
Prophet provided by the Python packages. The architectures of DeepAR, TCN and ConvTrans 
are as follows. The number of layers and hidden units are $1$ and $16$ for 
DeepAR. The number of multi-heads and kernel size are $2$ and $3$ for 
ConvTrans. The kernel size is $3$ for TCN with dilations in $[1,2,4,8]$. 
We also set batch size as $128$ and the number of epochs as $500$ for all 
forecasting methods.


{\bfseries Experiment results.}
Following \cite{li2019enhancing, rangapuram2018deep, tran2021radflow}, we 
evaluate the experimental methods using SMAPE and $\rho$-quantile loss 
$R_{\rho}$\footnote{Detailed definition is in supplementary.} with $\rho \in (0,1)$. The SMAPE results of all combination 
of clustering and forecasting methods are given in Table~\ref{tab-realdatasmape}. 
Table \ref{tab-realdatarloss} shows the $R_{0.5}/R_{0.9}$-loss for DeepAR, 
TCN and ConvTrans which give probabilistic forecasts. All results are 
run in $5$ trials. The best performance is highlighted by bold character. 
We observe that MixSeq is superior to other three methods, suggesting that 
clustering microscopic time series by our model is able to improve the 
estimation of macroscopic time series. Meanwhile, Macro and MixARMA have comparable performance and are better than DTCR, which further demonstrates the effectiveness 
of our method, i.e., only proper clustering methods are conductive to 
macroscopic time series forecasting.

\begin{table}
  \caption{Comparisons on the microscopic time series clustering methods for macroscopic time series forecasting combined with five forecasting methods: testing SMAPE on three real-world datasets.}
  \label{tab-realdatasmape}
  \centering
  \begin{tabular}{llllll}
      \toprule
      & & Macro & DTCR & MixARMA & MixSeq \\
      \midrule
      & ARMA & 0.2739(0.0002) & 0.2735(0.0106) & 0.2736(0.0013) & {\bf 0.2733}(0.0012) \\
      & Prophet & 0.1904(0.0007) & {\bf 0.1738}(0.0137) & 0.1743(0.0037) & 0.1743(0.0026) \\
      Rossmann & DeepAR & 0.1026(0.0081) & 0.1626(0.0117) & 0.1143(0.0088) & {\bf 0.0975}(0.0013) \\
      & TCN & 0.1085(0.0155) & 0.1353(0.0254) & 0.1427(0.0180) & {\bf 0.1027}(0.0075) \\
      & ConvTrans & 0.1028(0.0091) & 0.1731(0.0225) & 0.1022(0.0041) & {\bf 0.0961}(0.0019) \\
      \midrule
      & ARMA & {\bf 0.0540}(0.0001) & 0.0544(0.0018) & 0.0541(0.0003) & 0.0543(0.0001) \\
      & Prophet & 0.0271(0.0003) & 0.0271(0.0003) & 0.0269(0.0002) & {\bf 0.0267}(0.0002) \\
      M5 & DeepAR & {\bf 0.0278}(0.0034) & 0.0410(0.0046) & 0.0319(0.0063) & 0.0298(0.0029) \\
      & TCN & 0.0412(0.0075) & 0.0447(0.0044) & 0.0395(0.0094) & {\bf 0.0358}(0.0014) \\
      & ConvTrans & 0.0274(0.0048) & 0.0253(0.0020) & 0.0245(0.0024) & {\bf 0.0227}(0.0006) \\
      \midrule
      & ARMA & {\bf 0.0362}(0.0001) & 0.0363(0.0006) & 0.0364(0.0005) & {\bf 0.0362}(0.0002) \\
      & Prophet & {\bf 0.0413}(0.0001) & 0.0423(0.0008) & 0.0434(0.0003) & 0.0420(0.0005) \\
      Wiki & DeepAR & 0.0481(0.0008) & 0.0552(0.0015) & 0.0489(0.0006) & {\bf 0.0470}(0.0002) \\
      & TCN & 0.0494(0.0076) & 0.0654(0.0022) & 0.0491(0.0015) & {\bf 0.0446}(0.0023) \\
      & ConvTrans & 0.0471(0.0029) & 0.0497(0.0012) & 0.0466(0.0001) & {\bf 0.0440}(0.0010) \\
      \bottomrule
  \end{tabular}
\end{table}

\subsection{Sensitivity analysis of cluster number}
\label{exp-part3}
\begin{figure}
  \centering
  \includegraphics[width=1.0\linewidth]{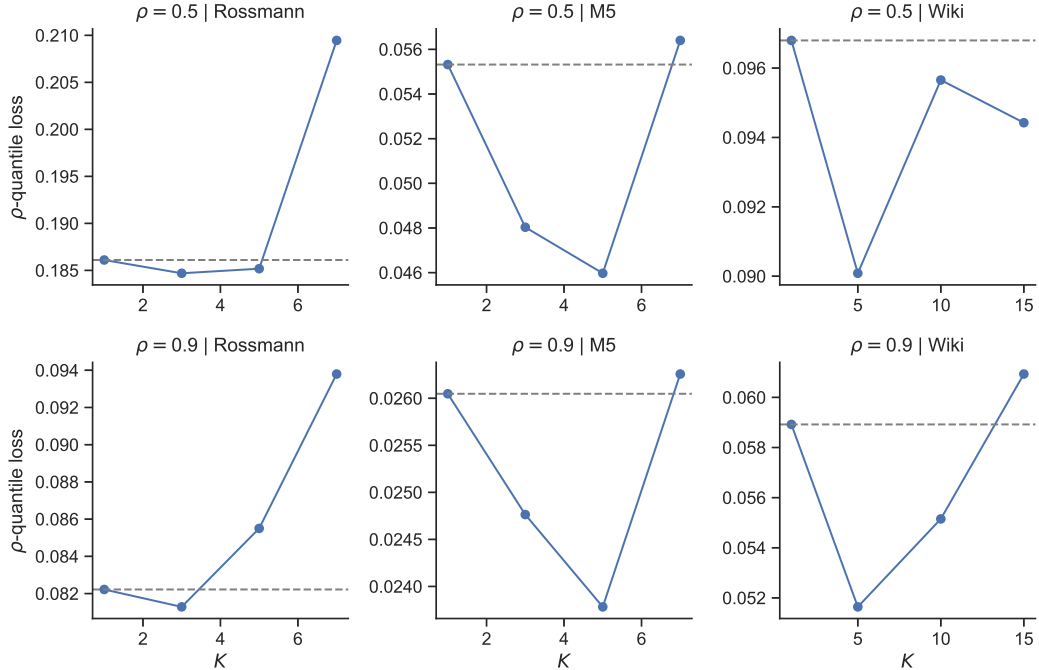}
  \caption{The macroscopic time series forecasting performance based on MixSeq with different cluster number $K$ on three real-world datasets. The time series forecasting method is fixed as causal convolution Transformer. Top three figures show the $R_{0.5}$-loss and bottom three figures show the $R_{0.9}$-loss.}
  \label{fig:diffk}
\end{figure}

The cluster number $K$ is a critical hyperparameter of MixSeq. To analyze
its effect on the forecasting of macroscopic time series, we conduct
experiments on both synthetic data and real-world data.
The results state the importance of setting a proper number of
clusters. We suggest do binary search on this critical hyperparameter.
Details are as follows.

{\bfseries Synthetic data.}
Following the experimental setting in section~\ref{exp-part1}, we generate $10,000$ 
microscopic time series from $3$ different parameterized ARMA respectively. That is, the
ground truth number of clusters is $3$ and there are $30,000$ microscopic samples in
total. The aggregation of all samples is the macroscopic time series which is the 
forecasting target of interest. Then, we compare the forecasting performance between
the method that directly forecasts macroscopic time series (denoted as Macro) and our
method with different cluster numbers (including $2$, $3$, $5$, denoted as MixSeq\_2,
MixSeq\_3 and MixSeq\_5 respectively). We fix the forecasting method as ARMA, and apply
the rolling window approach for T+10 forecasting in the last $40$ time steps. 
The average SMAPE of $5$ trials are $0.807$, $0.774$, $0.731$ and $0.756$ for Macro, MixSeq\_2, MixSeq\_3 and MixSeq\_5 
respectively. MixSeq\_3 being set with ground truth cluster number shows the best 
performance, while MixSeq\_2 and MixSeq\_5 would degenerate though still better than 
Macro. This result shows the importance of setting a proper number of clusters.

{\bfseries Real-world data.}
We do the evaluation on three real-world datasets by varying the cluster number $K$ of 
MixSeq while maintaining the other parameters fixed. For Rossmann and M5 datasets, we set 
the cluster number $K \in \{3,5,7\}$, while we explore the cluster number
$K \in \{5,10,15\}$ on Wiki dataset. The architecture and training parameters of MixSeq are same
to section~\ref{exp-part2}, except that we set the dropout rate as $0.1$, the penalty weight on the 
$\ell_2$-norm regularizer as 5e-5, and the learning rate as 1e-4. Meanwhile, 
we also fix the time series forecasting method as causal convolution Transformer (ConvTrans).

Figure~\ref{fig:diffk} reports the macroscopic time series forecasting performance (testing 
on $R_{0.5}$ and $R_{0.9}$ loss) based on MixSeq with different cluster number $K$ 
on three real-world datasets. The horizontal dashed lines are the results with $K=1$ 
that directly building ConvTrans model on the macroscopic time series without 
leveraging the microscopic data (named as ``Macro'' in section~\ref{exp-part2}). It is obvious that 
each dataset has its own suitable number of clusters, and our method is relatively sensitive to 
$K$, especially on the dataset with less microscopic time series, such as Rossmann. 
Similar to the accurately modeling of each microscopic time series, the larger cluster number 
$K$ of MixSeq also brings large variance to macroscopic time series forecasting, which 
degrades the performance of our method.

\section{Conclusion}
In this paper, we study the problem that whether macroscopic time series forecasting
can be improved by leveraging microscopic time series. Under mild assumption of 
mixture models, we show that appropriately clustering microscopic time series 
into groups is conductive to the forecasting of macroscopic time series. We 
propose MixSeq to cluster microscopic time series, where all the 
components come from a family of Seq2seq models parameterized with different 
parameters. We also propose an efficient stochastic auto-encoding variational 
Bayesian algorithm for the posterior inference and learning for MixSeq. 
Our experiments on both synthetic and real-world data suggest that MixSeq 
can capture the characteristics of time series in different groups and 
improve the forecasting performance of macroscopic time series. 

\begin{ack}
This work is supported by Ant Group.
\end{ack}

\appendix

\setcounter{proposition}{0}
\section{Proofs}
\begin{proposition}
Assuming the mixture model with probability density function $f(x)$, and corrsponding components $\left\{f_i(x)\right\}_{i=1}^K$ with constants $\left\{p_i\right\}_{i=1}^K$ ($\left\{p_i\right\}_{i=1}^K$ lie in a simplex), we have $f(x) = \sum_i p_i f_i(x)$.
In condition that $f(\cdot)$ and $\left\{f_i(\cdot)\right\}_{i=1}^K$ have first and second moments, i.e., $\mu^{(1)}$ and $\mu^{(2)}$ for $f(x)$, 
and $\left\{\mu_i^{(1)}\right\}_{i=1}^K$ and $\left\{\mu_i^{(2)}\right\}_{i=1}^K$ for components $\left\{f_i(x)\right\}_{i=1}^K$, we have:
\begin{align}
\sum_i p_i\cdot \mathrm{Var}(f_i) \leq \mathrm{Var}(f).
\end{align}
\end{proposition}

\begin{proof}
We prove the result based on the fact that we have for any moment $k$ that
\begin{equation}
\begin{aligned}
\mu^{(k)} = \mathbb{E}_f\left[x^k\right] = \sum_i p_i \mathbb{E}_{f_i}\left[x^k\right] = \sum_i p_i \mu_i^{(k)}.
\end{aligned}
\end{equation}
We then derive the variance of mixture as 
\begin{equation}
\begin{aligned}
\mathrm{Var}(f) &= \sum_i p_i \mu_i^{(2)} - \left( \sum_i p_i \mu_i^{(1)} \right)^2
= \sum_i p_i\left(\mathrm{Var}(f_i) + \left( \mu_i^{(1)} \right)^2 \right) - \left( \sum_i p_i \mu_i^{(1)} \right)^2\\
&= \sum_i p_i \mathrm{Var}(f_i) + \sum_i p_i \left( \mu_i^{(1)} \right)^2 - \left( \sum_i p_i \mu_i^{(1)} \right)^2.
\end{aligned}
\end{equation}
Since the squared function is convex, by Jensen's Inequality we immediately have 
$\sum_i p_i \left( \mu_i^{(1)} \right)^2 \geq \left( \sum_i p_i \mu_i^{(1)} \right)^2$.
\end{proof}

\section{Complexity analysis and running time}
{\bfseries Complexity analysis of MixSeq.}
The complexity of MixSeq depends on the number of clusters $K$ and three network architectures, 
including convolution, multi-head self-attention and MLP. For time series
$x \in \mathbb{R}^{t\times d}$, where $t$ and $d$ are the length and dimension of 
time series respectively, the FLOPs (floating point operations) of convolution is $O(w_k\cdot d\cdot t\cdot d_k)$, 
where $w_k$ and $d_k$ are the size and number of convolution kernel. The FLOPs of multi-head 
self-attention is $O(h\cdot t^2\cdot d_k)$, where $h$ is the number of multi-heads. The FLOPs of MLP is 
$O(h\cdot t\cdot d_k^2)$. Finally, the FLOPs of MixSeq is $O(K(w_k\cdot d\cdot t\cdot d_k+h\cdot t^2\cdot d_k+h\cdot t\cdot d_k^2))$. 
Since $w_k$, $d_k$ and $d$ are usually smaller than $t$, so the time complexity can be 
simplified as $O(K\cdot h\cdot t^2\cdot d_k)$ which is similar to Transformer.
Time series data is always recorded by day or hour. There are only one thousand values even for 
the data recorded in three years, so our method is capable for dealing with them. 
Furthermore, some existing methods can also be used in MixSeq to accelerate the
computation of self-attention.

\begin{figure}
  \centering
  \includegraphics[width=1.0\linewidth]{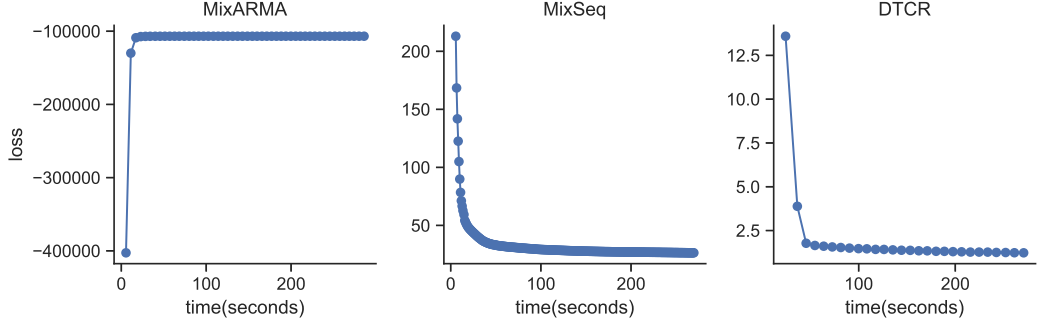}
  \caption{The convergence of different clustering methods over time (seconds) on Rossmann dataset. 
  The optimization objective of MixARMA is maximized by EM algorithm, 
  and the optimization objective of MixSeq and DTCR are minimized by gradient descent with Adam.}
  \label{fig:conv}
\end{figure}

{\bfseries Running time of macroscopic time series forecasting.}
The overall running time of forecasting macroscopic data based on MixSeq is comprised of two steps.
\textbf{(1)} The first step is to do clustering with MixSeq. Figure~\ref{fig:conv} shows
the convergence of MixSeq over time (seconds) compared with comparison approaches to
time series clustering. Our approach takes $200$ seconds to convergence on the dataset
containing $1,115$ microscopic time series, while MixARMA and DTCR take $20$ and $200$ seconds
to convergence respectively. The convergence rate of our method is not worse than existing neural network based approach, i.e., DTCR. \textbf{(2)} The second step is to forecast clustered time series with any proper forecasting model. The time complexity is in linear w.r.t the number of clustered time series. We can always accelerate this step by using more workers in parallel.

\section{Evaluation metrics}
\subsection{Rand index}
Given the labels as a clustering ground truth, Rand index (RI) measures the clustering accuracy between ground truth and predicted clusters, defined as
\[ \mathrm{RI} = \frac{a+b}{C_m^2}, \]
where $m$ is the total number of samples, $a$ is the number of sample pairs that are in the same cluster with same label, and $b$ is the number of sample pairs that in different clusters with different labels.

\subsection{Symmetric mean absolute percentage error}
Symmetric mean absolute percentage error (SMAPE) is an accuracy measure for time series forecasting based on percentage (or relative) errors. $ \mathrm{SMAPE} \in [0,1]$ is defined as
\begin{equation*}
  \mathrm{SMAPE} = \frac{1}{n} \sum_{t=1}^{n} \frac{\left|x_t - \hat{x}_t\right|}{\left|x_t\right|+\left|\hat{x}_t\right|},
\end{equation*}
where $x_t$ is the actual value and $\hat{x}_t$ is the predicted value for time $1 \le t \le n$, and $n$ is the horizon of time series forecasting.

\subsection{$\rho$-quantile loss}
In the experiments, we evaluate different methods by the rolling window 
strategy. The target value of macroscopic time series for each dataset is 
given as $x_{i,t}$, where $x_i$ is the $i$-th testing sample of macroscopic 
time series and $t\in [0,30)$ is the lead time after the forecast start 
point. For a given quantile $\rho \in (0,1)$, we denote the predicted 
$\rho$-quantile for $x_{i,t}$ as $\hat{x}_{i,t}^{\rho}$. To obtain such a 
quantile prediction from the estimation of clustered time series, a set of 
predicted samples of each clustered time series is first sampled. Then 
each realization is summed and the samples of these sums represent 
the estimated distribution for $x_{i,t}$. Finally, we can take the 
$\rho$-quantile from the empirical distribution.

The $\rho$-quantile loss is then defined as 
\begin{equation*}
  R_{\rho}(\mathbf{x}, \hat{\mathbf{x}}^{\rho}) = \frac{2 \sum_{i,t} D_{\rho} (x_{i,t}, \hat{x}_{i,t}^{\rho})}
  {\sum_{i,t} \left| x_{i,t} \right|}, \qquad
  D_{\rho} (x, \hat{x}^{\rho}) = (\rho - \mathbf{I}_{\{x \le \hat{x}^{\rho}\}})(x - \hat{x}^{\rho})
\end{equation*}
where $\mathbf{I}_{\{x \le \hat{x}^{\rho}\}}$ is an indicator function.

\section{Experiments}
\subsection{Enviroment setting}
We conduct the experiments on an internal cluster with $8$-core CPU, $32$G RAM and $1$ P100 GPU. 
Meanwhile, MixSeq, together with the time series forecasting methods based on neural network, 
are implemented with tensorflow 1.14.0.

\subsection{Simulation parameters of toy examples}
We use the TimeSynth\footnote{https://github.com/TimeSynth/TimeSynth} python package to 
generate simulation time series data. For GP time series, we use RBF as the kernel function. 
The lengthscale and variance are $[1.5,2]$, $[0.5,2.5]$ and $[0.5,1]$ for the mixture model 
of 3 GPs. Then, we add time series generated by GP with $[0.5,0.5]$ and $[2,1]$ for the 
samples from mixture of 5 GPs. Similarly, we use ARMA(2, 0) to generate ARMA time series. 
The parameters of the first three components of the mixture of ARMA are $[1.5,-0.75]$, $[1,-0.9]$ 
and $[-0.25,0.52]$ respectively. The parameters of another two components are $[0.34,0.27]$ 
and $[1,-0.30]$. The initial values of ARMA are sampled from $N(0,0.25)$.

\section{Societal impacts}
We study the problem that whether macroscopic time series forecasting
can be improved by leveraging microscopic time series, and finally propose MixSeq
to cluster microscopic time series to improve the forecasting of macroscopic time series.
This work will be especially useful for financial institutions and e-commercial platforms, 
e.g., loan forecasting, balance forecasting, and Gross Merchandise Volume (GMV) forecasting. 
The forecasting can help business decisions like controlling the risk of each financial institution, 
and help the lending to merchant in an e-commerce platform. 
Misuse of microscopic data could possibly lead to privacy issues. 
As such, protecting microscopic data with privacy-preserving techniques should be important.

\end{document}